\pdfoutput=1
\documentclass{bmvc2k}
\usepackage{multirow}
\usepackage{float}

\usepackage{amsmath,amsfonts,bm}

\def\eqref#1{equation~\ref{#1}}
\def\1{\bm{1}}

\def\eps{{\epsilon}}

\DeclareMathAlphabet{\mathsfit}{\encodingdefault}{\sfdefault}{m}{sl}
\SetMathAlphabet{\mathsfit}{bold}{\encodingdefault}{\sfdefault}{bx}{n}

\usepackage{hyperref}
\usepackage{url}
\usepackage{lipsum}
\usepackage{graphicx,caption}
\usepackage{graphicx}

\title{A Simple and Scalable Shape Representation for 3D Reconstruction}

\addauthor{Mateusz Michalkiewicz}{m.michalkiewicz@uq.net.au}{1}
\addauthor{Eugene Belilovsky}{eugene.belilovsky@umontreal.ca}{2}
\addauthor{Mahsa Baktashmotlagh}{m.baktashmotlagh@uq.edu.au}{1}
\addauthor{Anders Eriksson}{a.eriksson@uq.edu.au}{1}

 \addinstitution{University of Queensland \\
 Brisbane, Australia}

 \addinstitution{Mila, University of Montreal \\
 Montreal, Canada}

\runninghead{Michalkiewicz, Belilovsky, Baktashmotlagh, Eriksson}{eigenSDF}

\begin{document}

\maketitle

\begin{abstract}
Deep learning applied to the reconstruction of 3D shapes has seen growing interest. A popular approach to 3D reconstruction and generation in recent years has been the CNN encoder-decoder model usually applied in voxel space. However, this often scales very poorly with the resolution limiting the effectiveness of these models. Several sophisticated alternatives for decoding to 3D shapes have been proposed typically relying on complex deep learning architectures for the decoder model. In this work, we show that this additional complexity is not necessary, and that we can actually obtain high quality 3D reconstruction using a linear decoder, obtained from principal component analysis on the signed distance function (SDF) of the surface. This approach allows easily scaling to larger resolutions. We show in multiple experiments that our approach is competitive with state-of-the-art methods. It also allows the decoder to be fine-tuned on the target task using a loss designed specifically for SDF transforms, obtaining further gains. 
   
\end{abstract}

\section{Introduction}
\label{Sec:intro}
In recent years, we have witnessed an increased interest in extending the successes of deep learning to the analysis and representation of 3D shapes. This includes long standing problems, such as 3D shape reconstruction from single or multiple views~\citep{wu2016learning,choy20163d}, shape from silhouettes~\cite{cheung2003shape}, shape from contours~\cite{brady1984extremum}, and shape completion~\cite{mescheder2019occupancy}. 
Solutions to these problems can have a significant impact to applications in robotics~\cite{bylow2013real}, surgery~\cite{marmol2019dense}, 
and augmented reality~\cite{izadi2011kinectfusion}. 

One of the preferred categories of models for tackling these problems is the CNN encoder-decoder architecture \cite{choy20163d}, popularized originally in the context of segmentation~\citep{chen2017deeplab,long2015fully}. 
For example, in the single view reconstruction task a 2D CNN will encode the 2-D image and a 3D CNN decoder model will 
produce the final representation in voxels. Standard decoders, however, are ineffective in larger resolutions and do not make full use of the structure of the object.  
Similar problems arise in more general attempts to learn latent variable models of 3D shapes \citep{3dgan,gadelha20173d}. Here, one may be interested in tasks such as unconditional generation and reconstruction.

More recently authors have considered alternative representations of shapes to a standard 3D discretized set of voxels \citep{choy20163d, girdhar2016learning, wu2016learning, yan2016perspective, Zhu2017ICCV}, one that can permit more efficient learning and generation. These include point clouds \cite{fan2017point}, meshes~\citep{wang2018pixel2mesh, gkioxari2019mesh}, and signed distance transform based representations \citep{DeepLevelSets,park2019deepsdf}. To date there is not an agreed upon canonical 3-D shape representation for use with deep learning models nor a canonical decoder architecture for use with any of the described shape representations. Indeed, many complex alternative decoder architectures have been used \citep{richter2018matryoshka,tatarchenko2017octree}. In this work, we ask whether a very simple decoder architecture matched with the right shape representation can yield strong results. Building on the recent use of the Signed Distance Function (SDF) in shape representation we demonstrate a simple latent shape representation that can be used in downstream tasks and easily decoded. More specifically, in this work, we consider a latent shape representation obtained by applying PCA on the SDF transformed shape. We show this leads to a latent shape representation that can be used directly in downstream tasks like 3D shape reconstruction from a single view and 3D shape completion from a point cloud.

Our work 
a) reinforces the relevance of SDF as a representation for 3D deep learning; and
b) demonstrates that a simple representation obtained by applying PCA on the SDF transform can lead to an effective latent shape representation. This representation allows for results competitive to state of the art in standard benchmarks. Our work also suggests more complex benchmarks than the current ones may be needed to push forward the study of learned 3D shape reconstruction. 

The paper is structured as follows. In Sec.~\ref{sec:rel} we discuss the related work. 
We outline the basic methods used in the experiments in Sec.~\ref{sec:meth}. 
We show extensive quantitative and experimental results comparing our approach to existing methods 
in Sec.~\ref{sec:exp}.

\section{Related Work}\label{sec:rel} 
Several shape representations have been studied in the literature. Point cloud based representation requires a tedious step of sampling points from the surface and to generate shape subsequently inferring the continuous shape from a sample of points. Meshes present a challenge in that no clear way to generate valid meshes is available. Proposals have consisted of starting with template shapes and progressively deforming them \citet{wang2018pixel2mesh}. This, however, can be problematic as it never explicitly represents the shape and may suffer issues with local coherence.

Deep Level Sets \cite{DeepLevelSets} and DeepSDF \cite{park2019deepsdf} also use the SDF representation as in our work. Unlike our method, Deep Level Sets still relies on an encoder-decoder CNN architecture thus not removing the desired computational constraints associated with the 3D shape modeling. DeepSDF attempts to directly fit a continuous function to each shape which gives the SDF representations. Despite avoiding discretization, this function can lead to a complex decoder model, e.g. an 8 layer network is used to fit the SDF.  Another recent work \cite{mescheder2019occupancy}, similar in spirit to \citet{park2019deepsdf}, learns a classifier to predict whether a point is inside or outside of the boundary, using this classifier as the shape representation. Different from our proposal, these methods cannot easily learn a latent shape representation to be applied on downstream task, since the shape is represented by the weights of the classifier or regression model. On the other hand, our latent representation can easily encode an unseen shape and be conveniently used as a prediction target for deep learning models.

Our work can also be seen as complementary to the very recent observations in Tatar\-chenko et al. \cite{tatarchenko2019single} which highlights that good 3D single view reconstruction performance can be achieved by using retrieval or clustering methods. We note, however, that the descriptors used in that work are more complex.   

PCA has been classically used to represent shapes in a variety of contexts. For example, classical methods in computer vision such as the active appearance model \citet{edwards1998interpreting} and the 3D morphable model  used in face analysis \citet{blanz1999morphable} are based on PCA shape representations. However, these typically are applied in a different context requiring transforming the shape to a reference set of points and applying PCA on the coordinates. \citet{leventon2002statistical} used signed distance functions to embed 2D curves applying PCA to obtain statistical models. To the best of our knowledge it has not been combined with the SDF representing a surface in 3D. We note that level set methods and the SDF have only recently been revisited as an effective representation that can be combined with 3D deep learning \cite{DeepLevelSets,park2019deepsdf, chen2019learning}. Moreover, it is enlightening that this classic approach to shape representation can be competitive with deep learning methods on standard benchmarks.

\section{Methods}\label{sec:meth}
In this section, we start with reviewing the SDF transform and then describe our simple yet effective approach to shape representation.

\subsection{Signed Distance Functions}

Consider a 3D shape and its closed surface $\Gamma \subset \mathbb{R}^3$.
The \emph{Signed Distance Function} (SDF) of $\Gamma$ is a mapping 
$\phi:\mathbb{R}^3 \mapsto \mathbb{R}$ from any point 
$x \in \mathbb{R}^3$ to the surface:
  
\begin{equation}
    \phi(x) = \pm \inf_{y\in \Gamma} \Vert x-y \Vert, 
\end{equation}
with the convention that $\phi(x)$ is positive on the interior and 
negative on the exterior of $\Gamma$. 

In \citet{DeepLevelSets} a CNN decoder model is used to predict the SDF representation from a latent space as well as to learn autoencoders. We note, however, that this representation is well structured and objects are often grouped by category, we thus ask if a much simpler linear and non-convolutional decoder model can be effective at capturing its variability, leading to the \textit{eigenSDF} representation described in the next section.   

The above paper \cite{DeepLevelSets} further considers a loss function for the SDF representation that approximately minimizes the point-wise distance:
\begin{align}
&    L_\epsilon(\theta) = 
    \left( \sum_{x\in\Omega} \delta_\epsilon(\tilde \phi^j (x)) d^j(x)^p \right)^{1/p}  
    +\alpha \sum_{x\in\Omega} (\|\nabla \tilde \phi ^ j (x)\|-1)^2
\label{eq:lossfunctioneps}
\end{align}
with $\theta$ being parameters of the network, $\alpha$ a weighting factor, $\Omega$ an equidistant grid on which $\phi$ is evaluated, $\delta_\eps$ approximated Dirac delta, $\tilde \phi$ inferred Signed Distance Function, and $d(x)$ the closest distance between grid point $x$ and the ground truth shape. We will use this loss function to fine-tune our decoder model in the sequel.

\begin{figure}[t]

\includegraphics[width=0.99\textwidth]{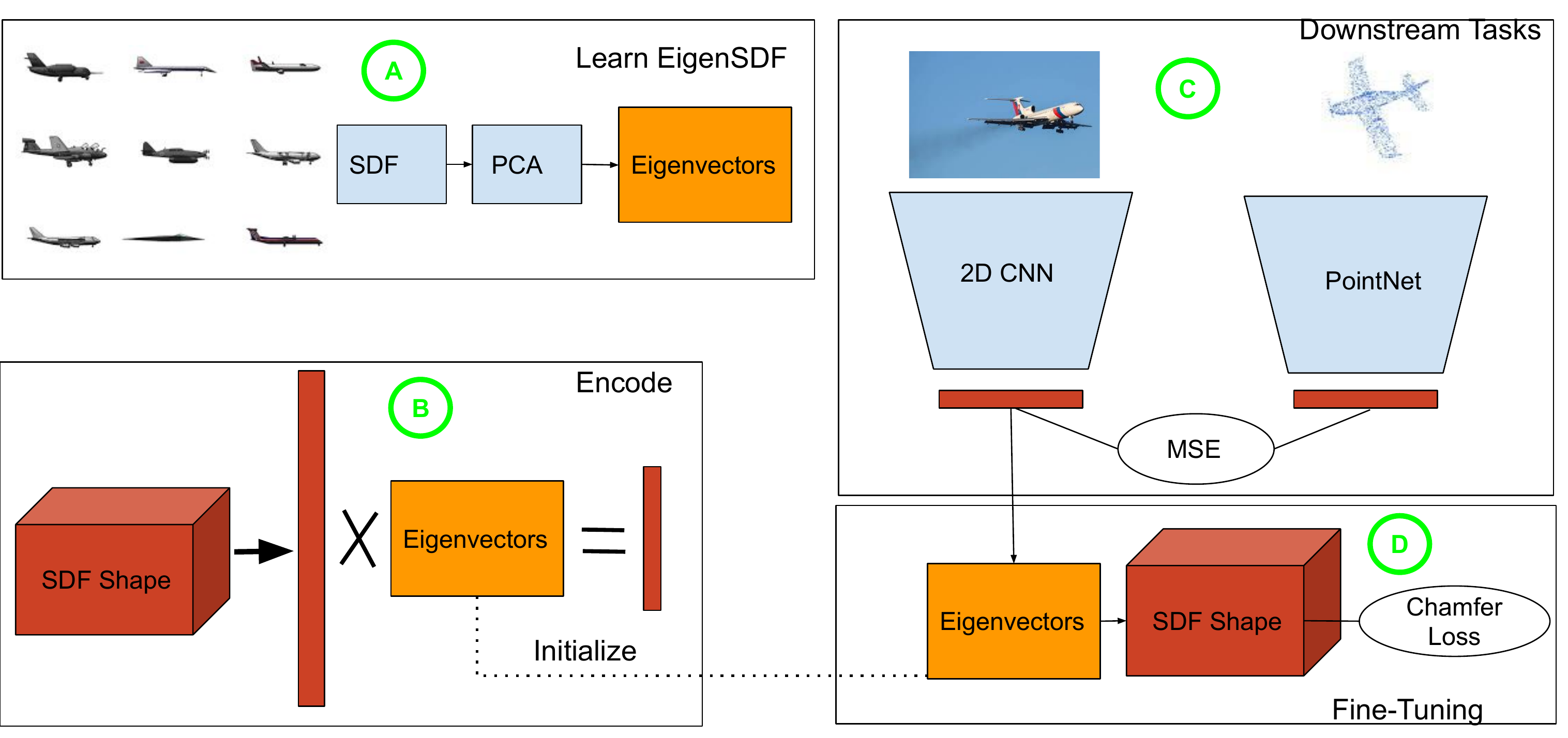}
\setlength{\abovecaptionskip}{5pt plus 3pt minus 2pt}
\caption{Overview of our experiments. We first apply PCA to all ShapeNet categories in order to retrieve eigenvectors (A). We encode every shape by applying eigenvectors to the signed distance function that is representing it (B). Our network for various experiments, \textit{eigenSDF}, consists of an input encoder (2D CNN or PointNet) and a linear layer (C). It uses an $\ell_2$ loss between its output and shape encodings from (B). We can directly decode predictions from the 2D CNN or PointNet using the eigenvectors. We can also further enhance performance, \textit{eigenSDF (finetuned)}, by finetuning the eigenvectors with the loss function in Eq.~\ref{eq:lossfunctioneps} that is designed for SDF representation (D). Here, the weights of the decoder are initialized from eigenvectors obtained in (B).
}\label{fig:main_fig}
\end{figure}

\subsection{EigenSDF}
We apply the PCA transform to $\mathbf{\phi_\text{all}}=\{\phi_i\}_{i=1..N}$, with $N$ being the number of training examples. The eigenvectors $E$ have the shape of $(k, M^3)$ with $M$ being the grid resolution and $k$ being the number of used eigenvectors. We project each SDF $\phi$ to the latent representation $\phi_c$ using the eigenvectors $E$: $\phi_c = \phi E^T$. Here, $\phi_c$ has a shape of $(1, k)$. In the sequel, we will denote this representation as the \textit{eigenSDF}. Note that applying PCA to the naive voxel representation would be inappropriate as the data is binary and therefore ill-suited for linear subspace methods such as PCA. 

For downstream tasks we predict directly the latent representation $\phi_c$. We will also consider using $E$ as an initialization which is finetuned by training on the SDF shape representation directly using Eq.~\ref{eq:lossfunctioneps}. A high level overview of our framework is given in Figure~\ref{fig:main_fig}.

\section{Experiments}\label{sec:exp}

We evaluate the proposed representations on 3 tasks: 
i) 3D reconstruction; 
ii) 3D reconstruction from point cloud; and 
iii) 3D reconstruction with autoencoders.  

These applications are evaluated on 13 categories from the ShapeNet repository \citep{chang2015shapenet}. 
\paragraph{Preprocessing.} In order to work on SDFs, we need to have a well defined interior and exterior of an object. 
We first preprocess the meshes to make them watertight using the method proposed in \cite{Stutz2018CVPR}. 
Following common practice, we render each ground truth mesh into 24 2D views using equally spaced azimuth angles. 
For each ground truth mesh, we compute a corresponding SDF in a $128\times128\times128$ discretized voxel grid.

\paragraph{Metrics.} Following the \cite{mescheder2019occupancy} experimental setup, we report 3 metrics. 
The first one is Intersection over Union (IoU), also known as Jaccard Index, between ground truth shape $S$ and prediction $\tilde S$:

$$ \text{IoU} = \frac{\vert S \cap \tilde S \vert}{\vert S \cup \tilde S \vert}.$$
The second metric measures point-wise distance between ground truth point set $S_P$ and prediction $\tilde S_Q$ using the symmetric Chamfer distance:
$$\text{chamfer}(S_P, \tilde S_Q) = \frac{1}{2\vert P \vert}\sum_{p\in P}\min_{q\in Q}\vert p-q \vert+\frac{1}{2\vert Q \vert}\sum_{q\in Q}\min_{p\in P}\vert p-q \vert.$$
Finally, we measure the angular distance using normal consistency (nc) metric:
$$\text{nc}(S_P, \tilde S_Q) = \frac{1}{2\vert P\vert} \sum_{p \in P}\vert N_{S_P}(p)\cdot N_{\tilde S_Q}(n_{\tilde S_Q}(p)) \vert + \frac{1}{2\vert Q\vert} \sum_{q \in Q}\vert N_{\tilde S_Q}(q)\cdot N_{S_P}(n_{S_P}(q)) \vert,$$
where $N_S(p)$ denotes normal of point $p$ lying on surface $S$ and $n_S(q)$ denotes nearest neighbour of point $q$ lying on surface $S$.

\subsection{3D Reconstruction from Single 2D View}\label{3dr}
In this set of experiments, we evaluate the \textit{eigenSDF} approach described in Sec~\ref{sec:meth}. We perform PCA jointly on all categories using a starting resolution of $128\times128\times128$. For memory efficiency, we use incremental PCA \cite{ross2008incremental}. $k$ eigenvectors were chosen to capture at least 99.5\% of the variance within the dataset. The image encoder is a 2D CNN whose architecture is taken from \cite{richter2018matryoshka}. We minimize the $\ell_2$ loss between the SDF projected into the latent space $\phi_c$, and the prediction of the 2D CNN. This network is trained for 100 epochs using an ADAM \cite{kingma2014adam} optimizer. Initial learning rate was set to $10^{-3}$ and dropped at epoch 30 to $10^{-4}$. Furthermore, we consider finetuning the representation starting with the eigenvectors from PCA and using Eq~\ref{eq:lossfunctioneps}. This baseline is referred to as \textit{eigenSDF (finetuned)}. 

In order to demonstrate that a gain is made by PCA versus just architecture, we also train a linear autoencoder of the same size ($M\times k$) and finetune it with Eq~\ref{eq:lossfunctioneps}. This baseline is referred to as \textit{linearSDF} and \textit{linearSDF(finetuned)}.

Finally, we compare to a set of standard benchmarks from the recent literature including voxel based CNN encoder-decoder \cite{choy20163d}, point cloud based methods \cite{fan2017point}, a mesh based method \cite{wang2018pixel2mesh}, and the recently introduced ONet \cite{mescheder2019occupancy}.

Complete results are given in Table \ref{tab:single_view}. First, we observe that simply using a same sized linear model \textit{linearSDF (finetuned)} is outperformed by using the \textit{eigenSDF}. Compared to alternatives, our method gives more significant gains in Chamfer metric than all competitors and can be further improved with the finetuning. We also observe improvements in the normal consistency metric. For the IoU metric, we observe that \textit{eigenSDF} outperforms all methods except \citet{mescheder2019occupancy}. Note 
that according to \citet{pix3d} Chamfer distance is a far better metric for shape comparison than IoU.

\vspace{-0.90cm}
\begin{table}[H]
\setlength{\abovecaptionskip}{5pt plus 3pt minus 2pt}

\centering
\resizebox{\textwidth}{!}{

\begin{tabular}{ |l|c|c|c|c|c|c| c|c|c|c| }
 \hline

 \multicolumn{10}{|c|}{Chamfer$\downarrow$} \\
 \hline
 Cat name   & 3D R2N2 & PSGN   & Pix2Mesh & AtlasNet  & ONet      &  linearSDF &  linearSDF (ft) & eigenSDF~  & eigenSDF (ft)~\\
 \hline
airplane    &  0.227  & 0.137  & 0.187    & 0.104     & 0.147     &   0.262    & 0.253           &   0.093    & \textbf{0.078}   \\
bench       &  0.194  & 0.181  &  0.201   & 0.138     & 0.155     &   0.255    &   0.243         &     0.091  &  \textbf{0.076}    \\
cabinet     &  0.217  & 0.215  & 0.196    & 0.175     & 0.167     &  0.229     &   0.222         &   0.077    &  \textbf{0.062} \\
car         &  0.213  & 0.169  & 0.180    & 0.141     & 0.159     & 0.233      & 0.230           &  0.068     &  \textbf{0.055}  \\
chair       &  0.270  & 0.247  & 0.265    & 0.209     & 0.228     &  0.269     &   0.262         &  0.113     &    \textbf{0.095}  \\
display     &   0.314 & 0.284  & 0.239    & 0.198     & 0.278     & 0.285      & 0.278           &  0.112     &  \textbf{0.110}    \\
lamp        &   0.778 & 0.314  & 0.308    & 0.305     & 0.479     &  0.642     &  0.627          &    0.469   & \textbf{0.388} \\ 
loudspeaker &   0.318 &  0.316 & 0.285    & 0.245     & 0.300     &  0.261     &  0.255          &   0.101    & \textbf{0.095}    \\
rifle       &   0.183 & 0.134  & 0.164    & 0.115     & 0.141     &    0.291   & 0.271           & 0.141      &   \textbf{0.139}\\
sofa        &   0.229 & 0.224  & 0.212    & 0.177     & 0.194     &  0.275     & 0.268           &   0.192    & \textbf{0.137} \\   
table       &   0.239 & 0.222  & 0.218    & 0.190     &  0.189    &  0.207     &  0.196          &   0.124    &  \textbf{0.111}     \\
telephone   &   0.195 & 0.161  & 0.149    & 0.128     & 0.140     &   0.175    &  0.170          &   0.051    &  \textbf{0.047}\\
vessel      &   0.238 & 0.188  & 0.212    & 0.151     & 0.218     &   0.425    &  0.414          &     0.351  & \textbf{0.339}   \\
 \hline
 mean       &   0.278  & 0.215  & 0.216   &  0.175    &  0.215    & 0.292      &0.283            & 0.152      & \textbf{0.133} \\
 \hline
 
\end{tabular}
}
\centering
\resizebox{\textwidth}{!}{
\begin{tabular}{ |l|c|c|c|c|c|c| c|c|c|c| }
 \hline
 \multicolumn{10}{|c|}{IoU$\uparrow$} \\
 \hline
  Cat name   & 3D R2N2 & PSGN  & Pix2Mesh         & AtlasNet  & ONet              &  linearSDF &  linearSDF (ft) & eigenSDF~& eigenSDF (ft)~ \\
 \hline
 airplane   &  0.426  & -     &   0.420          & -         &   \textbf{0.571}  & 0.421      & 0.432           & 0.524    &  0.541   \\
 bench      &  0.373  & -     &   0.323          & -         &   \textbf{0.485}  & 0.368      & 0.378           & 0.372    &  0.393   \\
 cabinet    &  0.667  & -     &   0.664          & -         &   \textbf{0.733}  & 0.655      & 0.667           & 0.688    &  0.703   \\
 car        &  0.661  & -     &   0.552          & -         &  \textbf{0.737}  & 0.666      & 0.679           & 0.716    & 0.732    \\
 chair      &  0.439  & -     &   0.396          & -         &   \textbf{0.501}  & 0.382      & 0.401           & 0.401    & 0.412      \\
 display    &  0.440  & -     &   \textbf{0.490} & -         &   0.471           & 0.385      & 0.397           & 0.431    & 0.439     \\
 lamp       &  0.281  & -     &   0.323          & -         &   \textbf{0.371}  & 0.208      & 0.215           & 0.234    & 0.275 \\
 loudspeaker&  0.611  & -     &   0.599          & -         &   \textbf{0.647}  & 0.558      & 0.566           & 0.596    & 0.606     \\
 rifle      &  0.375  & -     &   0.402          & -         &   \textbf{0.474}  & 0.259      & 0.265           & 0.392    & 0.395     \\
 sofa       &  0.626  & -     &   0.613          & -         &   \textbf{0.680}  & 0.606      & 0.621           & 0.624    & 0.639     \\
 table      &  0.420  &  -    &   0.395          &-          &   \textbf{0.506}  & 0.393      & 0.399           & 0.419    & 0.430     \\
 telephone  &  0.611  & -     &   0.661          & -         &   \textbf{0.720}  & 0.588      & 0.611           & 0.680    & 0.714 \\
 vessel     &  0.482  & -     &   0.397          & -         &   \textbf{0.530}  & 0.447      & 0.451           & 0.476    &  0.501  \\
 \hline
 mean       & 0.493   & -     &   0.480          & -         &   0.571           &  0.456     & 0.467           &  0.504   & 0.521  \\

\hline
 
\end{tabular}
}
\centering
\resizebox{\textwidth}{!}{

\begin{tabular}{ |l|c|c|c|c|c|c|c|c|c|c| }
 \hline
 \multicolumn{10}{|c|}{Normal Consistency$\uparrow$} \\
 \hline
 Cat name   & 3D R2N2  &PSGN& Pix2Mesh & AtlasNet & ONet           &  linearSDF &  linearSDF (ft) & eigenSDF~& eigenSDF (ft)~\\
 \hline
 airplane   &   0.629  & -  & 0.759    & 0.836    & \textbf{0.840} & 0.707      & 0.715           & 0.819    &  0.822         \\
 bench      &   0.678  & -  & 0.732    & 0.779    & 0.813          & 0.748      & 0.763           & 0.817    &   \textbf{0.828}   \\
 cabinet    &   0.782  & -  & 0.834    & 0.850    & 0.879          & 0.773      & 0.777           & 0.885    & \textbf{0.889}      \\
 car        &   0.714  & -  & 0.756    & 0.836    & 0.852          & 0.781      & 0.799           & 0.874    &  \textbf{0.878}        \\
 chair      &   0.663  & -  & 0.746    & 0.791    & 0.823          & 0.751      & 0.772           & 0.815    &   \textbf{0.827}     \\
 display    &   0.720  & -  & 0.830    & 0.858    & 0.854          & 0.750      & 0.781           & 0.870    & \textbf{0.877} \\
 lamp       &   0.560  & -  & 0.666    & 0.694    & 0.731          & 0.579      & 0.585           & 0.783    & \textbf{0.792}    \\
 loudspeaker&   0.711  & -  & 0.782    & 0.825    & 0.832          & 0.733      & 0.749           & 0.855    &    \textbf{0.862}  \\
 rifle      &   0.670  & -  & 0.718    & 0.725    & 0.766          & 0.661      & 0.669           & 0.816    & \textbf{0.819}  \\
 sofa       &   0.731  & -  & 0.820    & 0.840    & \textbf{0.863} & 0.738      & 0.740           & 0.855    &          0.861    \\
 table      &   0.732  & -  & 0.784    & 0.832    & \textbf{0.858} & 0.722      & 0.729           & 0.804    & 0.811          \\
 telephone  &   0.817  & -  &  0.907   & 0.923    & 0.935          & 0.830      & 0.852           & 0.921    & \textbf{0.936} \\
 vessel     &   0.629  & -  &  0.699   & 0.756    & 0.794          & 0.700      & 0.733           & 0.815    &  \textbf{0.817}   \\
 \hline
 mean       &   0.695  & -  & 0.772    &  0.811   & 0.834          & 0.728      & 0.743           & 0.840    & \textbf{0.847}          \\

 \hline

\end{tabular}
}
\caption{Single View 3D Reconstruction Results on ShapeNet. We observe that our \textit{eigenSDF} approach outperforms other state-of-the-art learning based methods in normal consistency and Chamfer distance. Finetuning can further improve this result. Compared to training a linear autoencoder or just finetuning the performance is substantially better, showing that eigen decomposition obtains the best results. }\label{tab:single_view}
\end{table}

\subsection{3D Shape Completion from Point Clouds}\label{shape_com}
We next consider shape completion from a point cloud. This task has been studied in \citep{liao2018deep,mescheder2019occupancy}.
Similar to experimental setup of \cite{mescheder2019occupancy}, we use 13 categories from ShapeNet repository and we pre-process the meshes to make them watertight. We randomly sample 300 points from ground truth meshes and add a Gaussian noise with 0 mean and 0.05 standard deviation. The same metrics have been used as described in section \ref{3dr}. 

We have encoded the input point cloud with PointNet encoder with a bottleneck dimension of 512 \cite{qi2017pointnet} and decoded it with linear decoder from section \ref{3dr}. A similar set of baselines has been used as in the previous section and compared to \textit{eigenSDF}. We observe similar large gains in the Chamfer metric and competitive performance in other metrics. 
Results in Table~\ref{exp:completion} show that, similar to 3D reconstruction task, our performance is much better in Chamfer distance, similar in normal consistency, and the second best in IoU.

\begin{table}[H]
\setlength{\abovecaptionskip}{5pt plus 3pt minus 2pt}

\centering
\begin{tabular}{ |l|c|c|c|  }

 \hline
 method & IoU$\uparrow$ &    Chamfer $\downarrow$    & nc $\uparrow$  \\
 \hline
 eigenSDF (ours)      & 0.568  &  \textbf{0.077}  & 0.852                   \\
 3D-R2N2 (\cite{choy20163d})       & 0.565 & 0.169 & 0.719                                  \\
PSGN (\cite{fan2017point})       & - & 0.144 & -                                            \\
DMC (\cite{liao2018deep})   &   0.674 & 0.117 & 0.848                                   \\
ONet (\cite{mescheder2019occupancy}) &    \textbf{0.778} & 0.079 & \textbf{0.895}           \\
 
 \hline
\end{tabular}
\caption{Results on 3D shape completion}\label{exp:completion}
\vspace{-2ex}
\end{table}

\subsection{3D Reconstruction from Latents}\label{ae3d}
Finally, we consider a simple 3D reconstruction task \cite{girdhar2016learning}. This can also be viewed as measuring the representational power of the model \cite{mescheder2019occupancy}.  We evaluate reconstruction quality of \textit{eigenSDF} versus other methods, particularly CNN-based autoencoders. The goal is to reconstruct test set shapes. An initial resolution of $128\times128\times128$ was used and reduced to $k=512$ as done in other works \cite{mescheder2019occupancy}. We use \textit{cars} category from ShapeNet repository and evaluate reconstruction on unseen data. For the evaluations, in addition to the metrics used in section \ref{3dr}, we further analyse the decoders using F-score \cite{Knapitsch2017}. Results are shown in Table~\ref{tab:ae}.

\begin{table}[H]
\setlength{\abovecaptionskip}{5pt plus 3pt minus 2pt}

\centering
\begin{tabular}{ |l|c|c|c|c|  }
 \hline
 method   & IoU$\uparrow$&   Chamfer$\downarrow$    & NC$\uparrow$ & F-score$\uparrow$ \\
 \hline
  eigenSDF   &  0.746   &    0.0425        & 0.869        & 0.484       \\
 eigenSDF (ft)      & \textbf{0.758}    &  \textbf{0.0325}  &\textbf{0.896} &\textbf{0.529}\\
 Linear ($\phi$) (chamfer)      &  0.582   &0.050            &0.773 &  0.315        \\
 Linear(voxels)      & 0.637 &     0.067            & 0.737 &  0.384        \\
 DLS (\cite{DeepLevelSets})     & 0.681    &   0.047            & 0.858&  0.103        \\
 TL (\cite{girdhar2016learning})  & 0.656    &   0.082                & 0.847&  0.081        \\ 
 \hline
\end{tabular}
\caption{We compare \textit{eigenSDF} to the state-of-the-art methods in terms of reconstruction. We find that \textit{eigenSDF} performs better than linear autoencoders trained on voxels or SDFs.}\label{tab:ae}
\vspace{-2ex}
\end{table}

\subsection{Comparison with Deep Level Sets}
In this section, we compare our method to the other recent approach relying on the signed distance transform \cite{DeepLevelSets} and learning with the chamfer loss $L_{\epsilon}$. This one, however, uses the CNN decoder model and does not learn a latent shape representation. We have chosen a similar experimental setup of 3 subsets each having 2 000 examples from ShapeNet repository: \textit{cars}, \textit{sofas}, \textit{chairs}. We observed that remaining 2 categories, \textit{bottles} and \textit{phones}, are too simple to allow for a difference in higher resolution. 

\begin{figure}[H]
\centering
\vspace{-0.8cm}
\includegraphics[width=0.6\textwidth, trim={0 0 0 -1.5cm}]{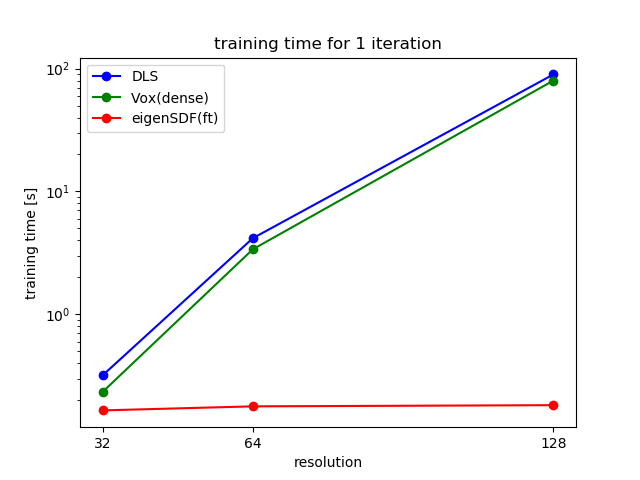}
\setlength{\abovecaptionskip}{0.25pt plus 0.3pt minus 0.2pt}
\caption{Training time of \textit{eigenSDF (finetuned)} and dense convnets (Voxels, DLS). Figure shows on a logscale amount of seconds a network needs for forward and backward pass of 1 iteration using a batchsize of 32. Resolution 256 not shown due to clarity.}\label{fig:training_time}
\end{figure}

We compare the training time for both methods for various resolutions. The results shown in Figure~\ref{fig:training_time} demonstrates that it is not feasible to use the CNN decoder in higher resolutions. It is consistent with the findings of \citep{richter2018matryoshka}. Reported metrics, which are shown in Table~\ref{tab:quantdls}, might differ slightly from \cite{DeepLevelSets} due to different pre-processing techniques. 
\vspace{-1cm}
\subsection{Reconstruction and Generation}
\begin{figure}[h]
\centering

\vspace{0.6cm}
\includegraphics[width=0.7\textwidth, trim={0 0 0 2cm}]{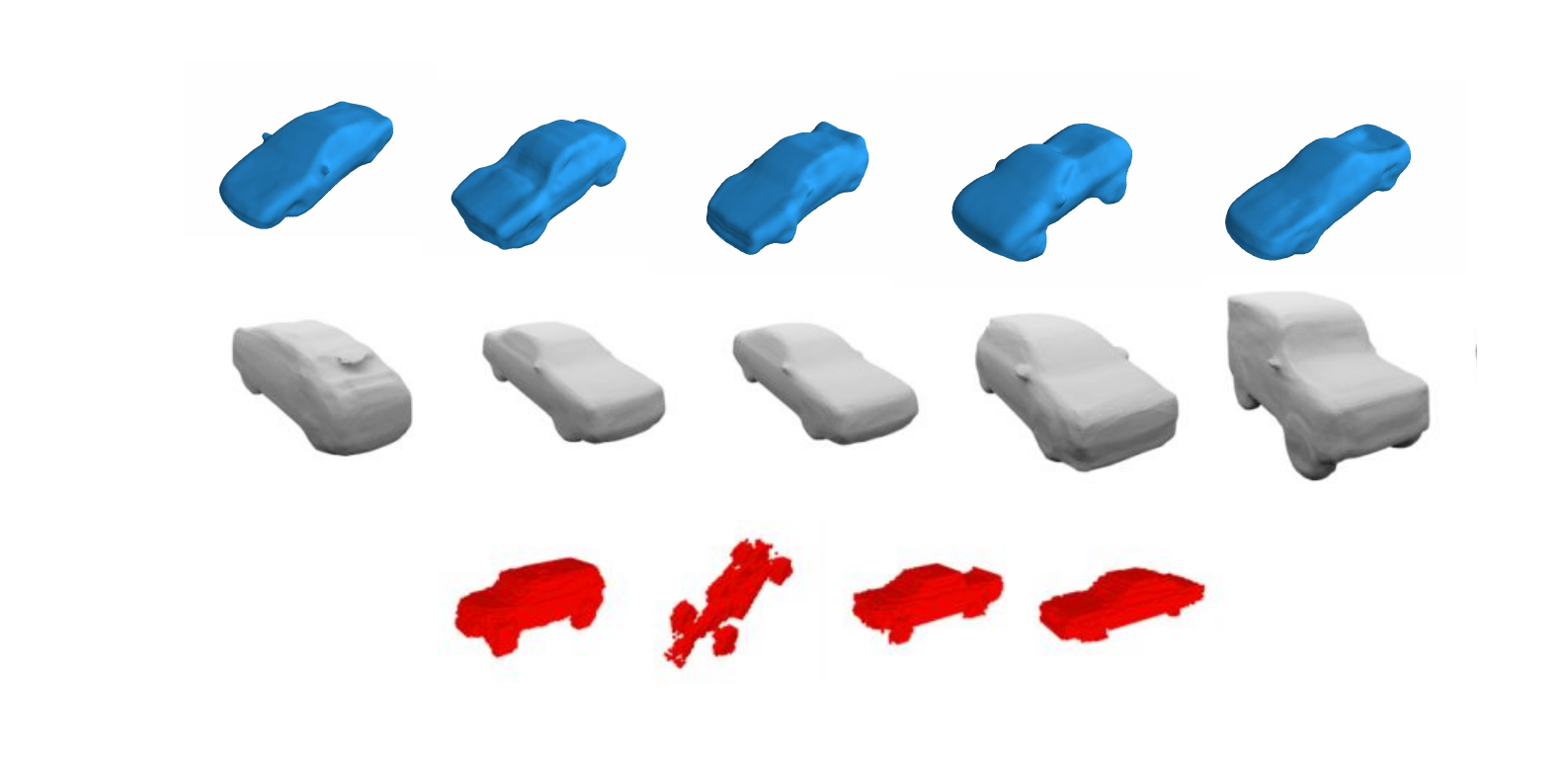}
\setlength{\abovecaptionskip}{-5.1pt}

\caption{We compare unconditional generations of \textit{cars} category. Generations from a gaussian fit to \textit{eigenSDF} is shown in the top row (blue). Second row are generations from \cite{mescheder2019occupancy} and the third row is from a 3D GAN \cite{3dgan}.}
\label{fig:unconditional}
\end{figure}

Finally, we evaluate the performance on the single view reconstruction qualitatively. In Figure~\ref{fig:qual_dls2}, we can see that reconstructions (on unseen data) can be effective capturing more complex structures ignored by \cite{DeepLevelSets}. Multiple authors have also consider generating unconditionally shapes, typically using sophisticated non-linear deep learning models like GANs and VAEs. We compare some of these to sampling a gaussian in the latent space of the \textit{eigenSDF}. Qualitative results are shown in Figure~\ref{fig:unconditional}. As it can be seen, our simple approach yields comparable shape representation to  the complex non-linear models.

\vspace{-1.0cm}
\section{Conclusion}
We have shown that using a simple linear decoder coupled with the SDF representation yields competitive results. The SDF lends itself effectively to the application of PCA yielding a strong but simple baseline for future work in learned 3D shape analysis. Moreover, our work suggests that more complex baseline datasets may be needed to further evaluate deep learning methods on 3D shape inference. 

\vspace{-0.5cm}
\paragraph{Acknowledgements.} 
Authors would like to thank Stavros Tsogkas for constructive comments. Eugene Belilovsky is funded by IVADO.

\begin{table}[th]
\begin{tabular}{p{0.5cm}p{2.5cm}p{2.5cm}p{2.5cm}p{2.5cm}  }

 & 
 \includegraphics[width=0.15\textwidth]{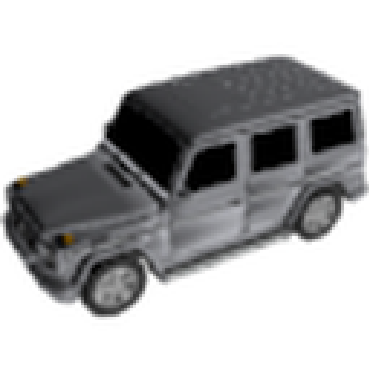}\hspace{1mm}   
 &  \includegraphics[width=0.15\textwidth]{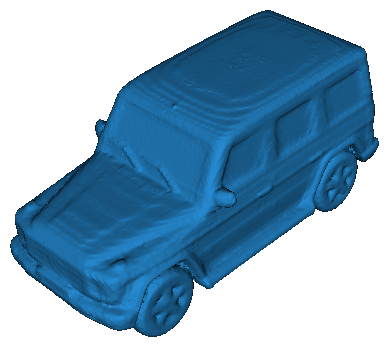}\hspace{1mm}         
 &\includegraphics[width=0.15\textwidth]{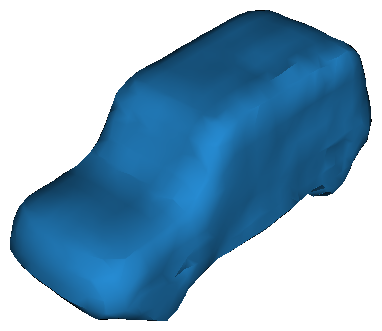}\hspace{1mm}
 &\includegraphics[width=0.15\textwidth]{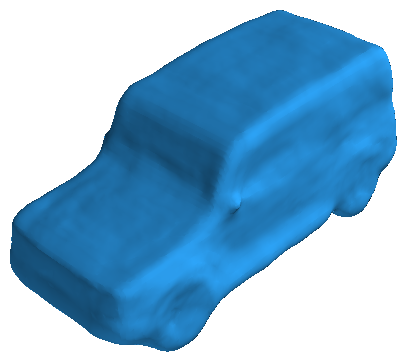}\hspace{1mm} \\

&
\includegraphics[width=0.13\textwidth]{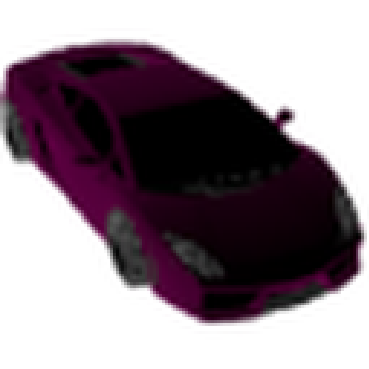}\hspace{1mm}   
 &  \includegraphics[width=0.15\textwidth]{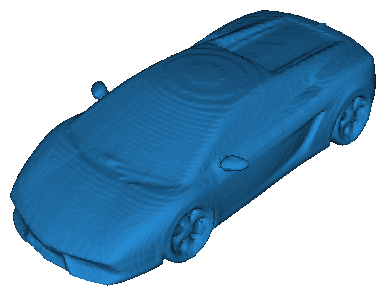}\hspace{1mm}         
 &\includegraphics[width=0.15\textwidth]{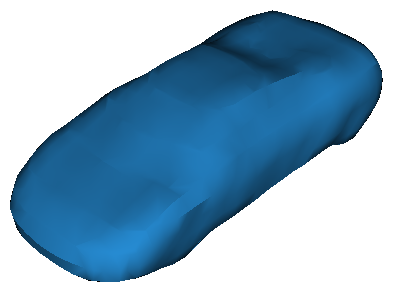}\hspace{1mm}
 &\includegraphics[width=0.15\textwidth]{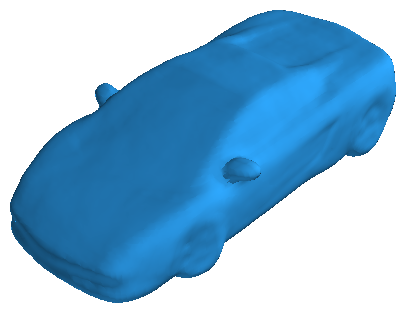}\hspace{1mm} \\

&
\includegraphics[width=0.19\textwidth]{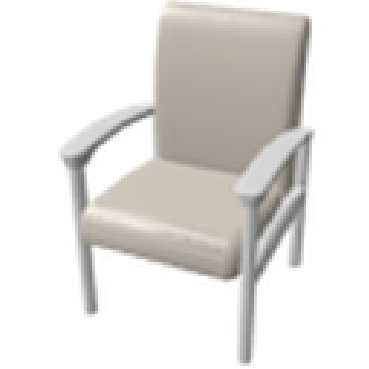}\hspace{1mm}  
 &  \includegraphics[width=0.15\textwidth]{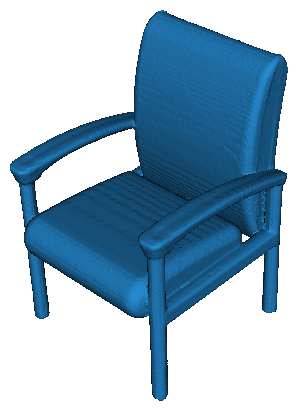}\hspace{1mm}    
 &\includegraphics[width=0.15\textwidth]{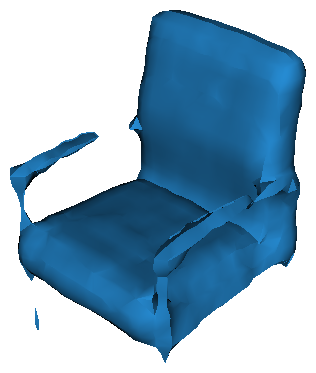}\hspace{1mm}
 &\includegraphics[width=0.15\textwidth]{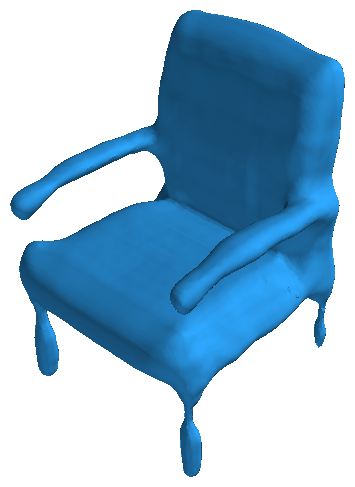}\hspace{1mm} \\

&
\includegraphics[width=0.20\textwidth]{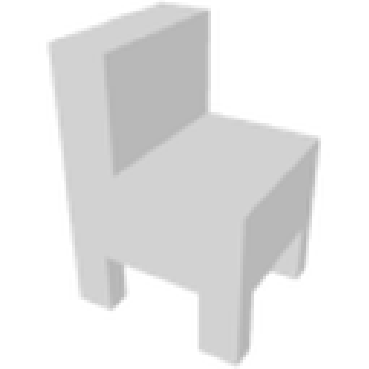}\hspace{1mm}  
 &  \includegraphics[width=0.15\textwidth]{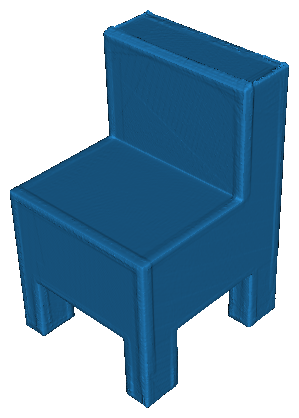}\hspace{1mm}    
 &\includegraphics[width=0.15\textwidth]{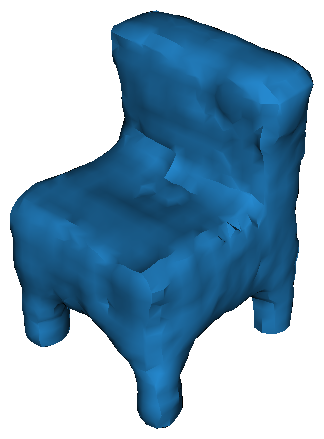}\hspace{1mm}
 &\includegraphics[width=0.15\textwidth]{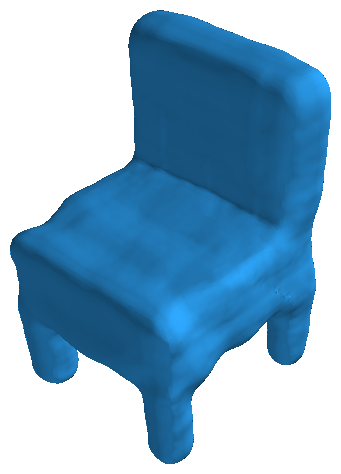}\hspace{1mm} \\

&
\includegraphics[width=0.13\textwidth]{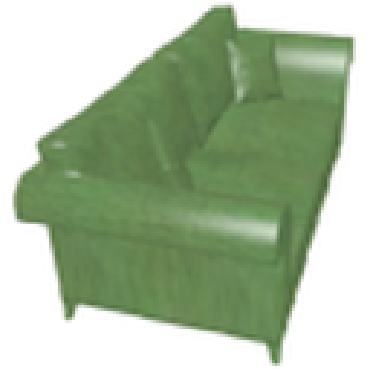}\hspace{1mm}   
 &  \includegraphics[width=0.15\textwidth]{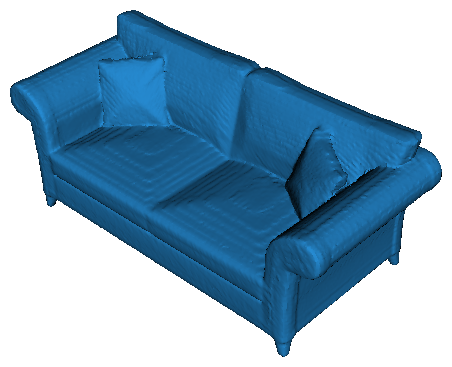}\hspace{1mm}     
 &\includegraphics[width=0.15\textwidth]{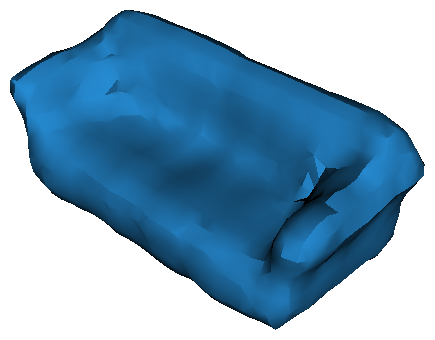}\hspace{1mm}
 &\includegraphics[width=0.15\textwidth]{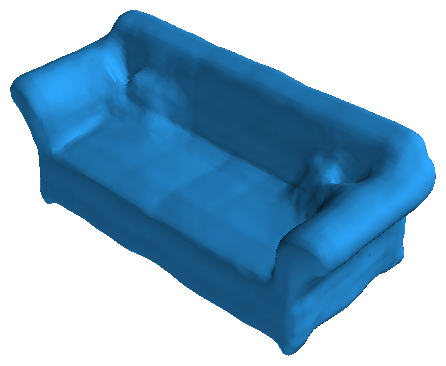}\hspace{1mm} \\

 &
\includegraphics[width=0.15\textwidth]{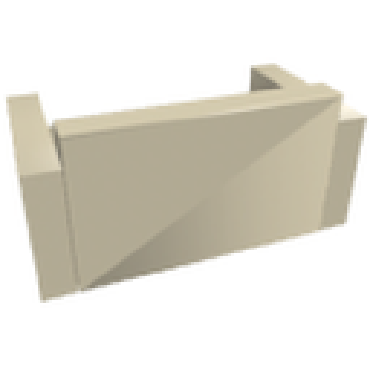}\hspace{1mm}   
 &  \includegraphics[width=0.15\textwidth]{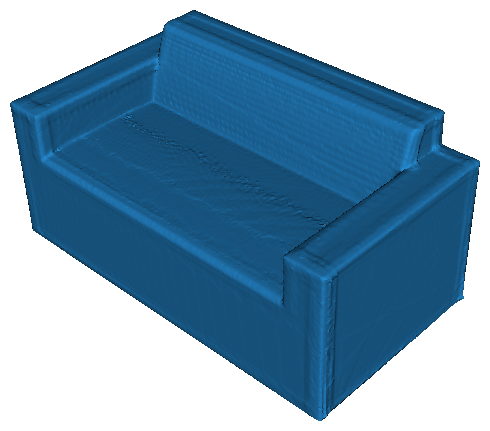}\hspace{1mm}     
 &\includegraphics[width=0.15\textwidth]{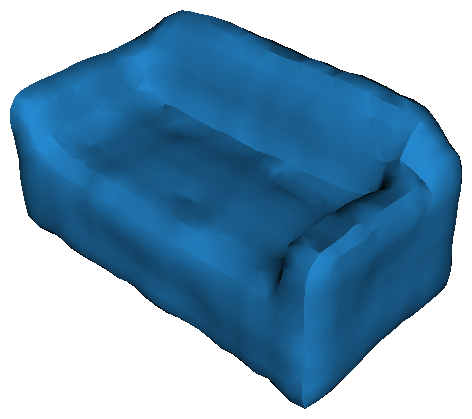}\hspace{1mm}
 &\includegraphics[width=0.15\textwidth]{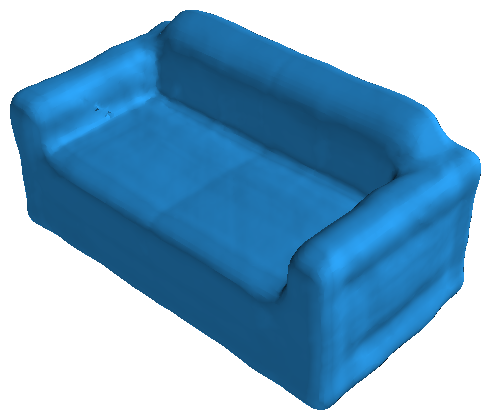}\hspace{1mm} \\
  &~~~~~2D view &  ~~~~~~~~GT    & Deep Level Sets & ~~eigenSDF 
\end{tabular}
\setlength{\abovecaptionskip}{5pt plus 3pt minus 2pt}

 \captionof{figure}{We compare reconstruction of \textit{eigenSDF} and the CNN decoder based Deep Level Sets \cite{DeepLevelSets}, which also uses SDF representation. \textit{eigenSDF} allows us to operate at a higher resolution and generally produces more locally coherent results.} \label{fig:qual_dls2}
\end{table}

\begin{table}[h]
\scriptsize
\centering
  \begin{tabular}{|l|l|l|l|l|l|l|l|l|l|l|l|l|}
    \hline

	category & 
      \multicolumn{4}{c}{DLS} &
      \multicolumn{4}{c|}{eigenSDF(ft)} \\
      \hline
    & IoU$\uparrow$ & Chamf$\downarrow$& NC$\uparrow$ & F-score$\uparrow$
    & IoU$\uparrow$ & Chamf$\downarrow$& NC$\uparrow$ & F-score$\uparrow$ \\
    \hline
    cars    & 0.784 & 0.055 & 0.804 & 0.148 & \textbf{0.821} & \textbf{0.040} & \textbf{0.909} & \textbf{0.432}  \\
    chairs  & 0.434 & 0.360 & 0.743 & 0.066 & \textbf{0.553} & \textbf{0.125} & \textbf{0.820} & \textbf{0.168}  \\
    sofas   & 0.581 & 0.132 & 0.779 & 0.089 & \textbf{0.647} & \textbf{0.082} & \textbf{0.867} & \textbf{0.248} \\
    
    \hline
  \end{tabular}
\setlength{\abovecaptionskip}{5pt plus 3pt minus 2pt}
  \caption{Comparison to the SDF based method \cite{DeepLevelSets} in single view reconstruction. There is marked improvement due to the ability to model higher resolution.} \label{tab:quantdls}
\end{table}

\clearpage

\bibliography{bmvc_arxiv}

\end{document}